\pdfoutput=1
\documentclass[11pt]{article}

\usepackage[]{EMNLP2022}

\usepackage{times}
\usepackage{latexsym}

\usepackage[T1]{fontenc}
\usepackage[utf8]{inputenc}

\usepackage{microtype}

\usepackage{inconsolata}

\usepackage{graphicx}
\usepackage{url}
\usepackage{latexsym}
\usepackage{makecell}
\usepackage{epstopdf}
\usepackage{tipa}
\usepackage{hyperref}
\usepackage{xstring}
\usepackage{amsfonts}
\usepackage{amsmath}
\usepackage{fixltx2e}
\usepackage[T1]{fontenc}
\usepackage{booktabs}
\usepackage{arabtex}
\usepackage{hhline}
\usepackage{pdfpages}
\usepackage{utf8}
\newcommand{\hide}[1]{}

\newcommand{\SHADDA}{{$\sim$}}

\usepackage{multirow}
\usepackage{subfigure}
\usepackage{amssymb}
\usepackage[super]{nth}
\usepackage{enumitem}
\usepackage{footmisc}
\title{Arabic Word-level Readability Visualization\\ for Assisted Text Simplification}

\author{Reem Hazim, Hind Saddiki, Bashar Alhafni, Muhamed Al Khalil, Nizar Habash \\
  Computational Approaches to Modeling Language Lab\\
  New York University Abu Dhabi\\
  \texttt{\{rh3015,hind.saddiki,alhafni,muhamed.alkhalil,nizar.habash\}@nyu.edu}
  }

\newcommand{\samer}{SAMER}

\setcode{utf8}
\vocalize

\begin{document}

\maketitle

%%% ADD SAMER project name...
\begin{abstract}
This demo paper presents a Google Docs add-on for automatic Arabic word-level readability  visualization. The add-on includes a lemmatization component that is connected to a five-level readability lexicon and Arabic WordNet-based substitution suggestions. The add-on can be used for assessing the reading difficulty of a text and identifying difficult words as part of the task of manual text simplification.  
%The add-on tool suite is accompanied with an offline API for optional superior, but more computationally expensive, morphological disambiguation.
We make our add-on and its code publicly available.\footnote{\url{http://samer-addon.camel-lab.com/}\label{demo}}$^,$\footnote{\url{https://github.com/CAMeL-Lab/samer-add-on}\label{api}} 
\end{list} % Don't remove, this is an issue with arabtex
\end{abstract}
%$^,$\footnote{Video: \url{http://samer-addon-vid.camel-lab.com/}}

% Notes
% \begin{itemize}
    
%     \item Arabic is a MRL and leveling is done at the lemma level, we need to have a morph analyzer
    
    % \item Why Google Docs? 1) ease of use; 2) popularity; 3) multi-editor; 4) offline process to generate the annotation for efficiency; 5) Light and fast
    
%     \item provide an easy way to integrate different readability levels (this should go in the design) and examples of this should be discussed in the implementation
    
% \end{itemize}
\section{Introduction}
Models for  automatic readability assessment and automatic text simplification are relevant to many natural
language processing (NLP) tasks such as developing pedagogical language technologies that assist students with learning languages, or teachers with curriculum design and writing assessment, as well as personalized paraphrasing of NLP systems' outputs to target different users with different readability levels.

\begin{figure}[t]
\centering
    \includegraphics[width=1.0\columnwidth]{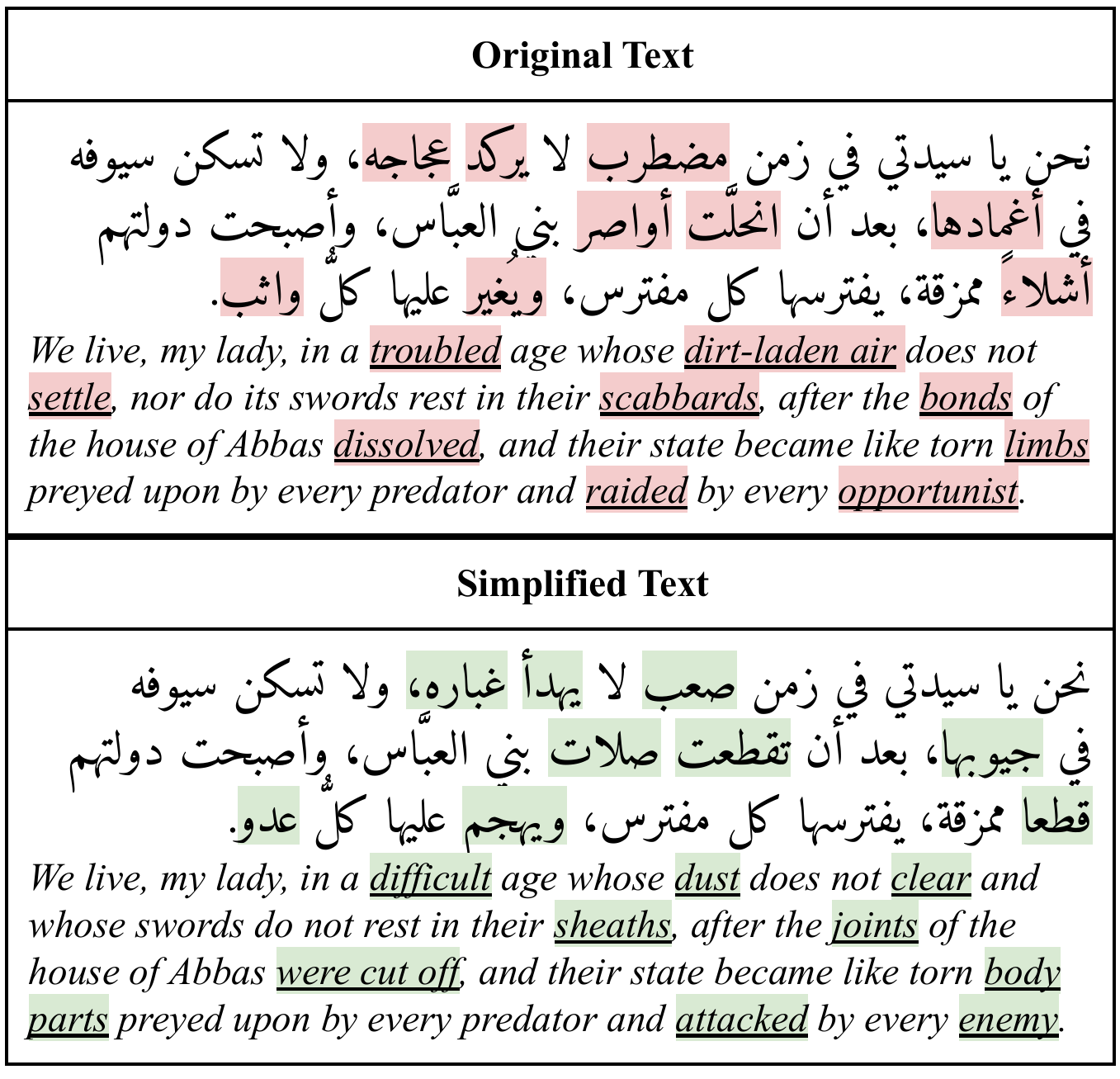}
\caption{ An example original sentence from the Arabic novel ``The Knight of Bani Hamdan'' \cite{aljarim:1945}. The red-marked words are all of readability level 4 and 5 (difficult) per \newcite{al-khalil-etal-2018-leveled}'s readability lexicon. The simplified text rewrites those words into lower (easier) levels (green-marked words).  The English translations are a best attempt to convey the complexity level of the Arabic word choices to  non-Arabic readers.}
\label{fig:intro}
\end{figure}

Developing robust models for readability assessment and simplification requires the creation of large-scale lexical and annotated resources for training and evaluation. 
For example,  parallel texts with different paired readability levels can be used to train readability models as well as simplification models. Figure~\ref{fig:intro} presents an example from an Arabic novel paired with a simplified version targeting the fifth-grade readability level.
Properly identifying which words and phrases need to be rewritten in a simplified manner for a specific target readability level and audience requires word-level readability annotation in a framework that enables easy editing of the original text, as well as easy checking on the updated text.  To our knowledge, most of the available tools for readability assessment work on the document or the sentence levels.
%and when they offer access to word level readability, they tend to be coarse. \textcolor{red}{Add citations or examples}. %add citations?

The system presented in this paper addresses this limitation by focusing on word-level readability visualization to assist human annotators working on identifying text readability levels and adjusting texts to simplify them in a controlled setting.  
%Naturally the same interface can be used as a language study assistant tool; or a as tool to identify appropriate reading materials for students of different learning levels.
%
This effort is part  of a project on the Simplification of Arabic Masterpieces for Extensive Reading
({\samer}) \cite{AlKhalil:2017:simplification,al-khalil-etal-2018-leveled,al-khalil-etal-2020-large,Jiang:2020:online}. The project goals include the creation of a lemma-based graded readability lexicon for Arabic and a corpus of parallel original and simplified texts from Arabic novels (such as those presented in Figure~\ref{fig:intro}). The project plans to target two different simplified readability levels: Grades 4-5 (Level III) and Grades 6-8 (Level IV).

While our focus is on Arabic, a language with limited annotated resources for text simplification,  the components we developed can be easily extended to other languages. 
The demo system is a Google Docs add-on that includes morphological analysis and light disambiguation of Arabic text, visualization of word readability levels, and access to substitution options with their own explicit readability levels. 
%Additionally we provide an offline API for optional superior, but more computationally expensive, morphological disambiguation.
We make our add-on and its code publicly available.\footref{demo}$^,$\footref{api}

Next, we present some relevant Arabic linguistic facts (\S\ref{sec:ling}), and discuss related work (\S\ref{related}).
We then present our design and implementation decisions  (\S\ref{design}). In \S\ref{example} we discuss some examples and use cases.

% What problem does the proposed system address?
% Why is the system important and what is its impact?
% What is the novelty in the approach/technology on which this system % is based?
% Who is the target audience?
% How does the system work?
% How does it compare with existing systems?
% How is the system licensed?

%Beyond diglossia and weak reading in MSA among children, there's also difficulty in curriculum building, targeted reading materials building, etc for educators who need this kind of help and content creators to ensure the work they're creating is both thematically and linguistically at the right level for the target audience

% do we need to keep this?
%\paragraph{Motivation}
%\begin{itemize}

%\item  Work has been done on readability in Arabic (cite papers from Hind). However, all the work there focuses on the document-level and not word-level. We should highlight word-level contributions!!

%\item   Work done by Ross and Khalil, focuses on word-level and not docuemnt.

%\item Our work focuses both at the document and word-level. Why is this important? MSA is the official language but not the spoken language daily. L1 and L2 learners have a challenge of reading text that is above their readability and writing text to navigate different readability levels for both the document and word level. Personalization/User-centric.

%\item We can also talk about this paper as part of a bigger project on text simplification. We are describing the tool with all its possible use cases. 
%Add the graphics to the beginning to motivate the paper.

%\end{itemize}

%\paragraph{Contributions}

\section{Relevant Arabic Linguistic Facts}
\label{sec:ling}

Modern Standard Arabic (MSA) poses many challenges for NLP tasks  \cite{Habash:2010:introduction}. Two in particular are directly relevant to the task at hand, and affect many of our design decisions:
morphological richness and orthographic ambiguity.\footnote{We do not handle dialectal variants in this effort, although we acknowledge that dialectal differences from MSA are an important factor in readability assessment, since MSA is not the native variant of Arabic learned at home \cite{Ferguson:1959:diglossia,Holes:2004:modern,Carroll:2017:triglossia}.}

\paragraph{Morphological Richness}
Arabic employs a combination of templatic, affixational, and cliticization morphological operations to realize a large number of features such as gender, number, person, case, state, aspect, voice, and mood, in addition to a number of attachable pronominal, preposition and determiner clitics.  This leads to a very large number of words to model. 
To address this aspect, we utilize a morphological analysis component that is optimized for efficient representation \cite{Graff:2009:standard,Taji:2018:arabic-morphological}.

%%% rewrite this next para
\paragraph{Orthographic Ambiguity}
Arabic is commonly written with optional diacritical marks -- which are often omitted -- leading to rampant ambiguity.
Orthographic ambiguity and morphological richness interact heavily with each other.
For example the word 
\<فردها>
{\it frdhA}\footnote{Arabic HSB transliteration \cite{Habash:2007:arabic-transliteration}.}
%\footnote{Arabic transliteration is presented in the Habash-Soudi-Buckwalter scheme \cite{Habash:2007:arabic-transliteration}.}
has four core lemmas \cite{Jiang:2020:online}: 
the verbs
\<فَرَّد>
{\it far{\SHADDA}ad}
`individualize, separate in units', and
\<رَدّ>~{\it rad{\SHADDA}}
`answer, return';
and the nouns
\<فَرْد>
{\it fard}
`individual, unit'
and
\<رَدّ>
{\it rad{\SHADDA}}
`response, return'.

\begin{table}[t]
\centering
\small
\setlength{\tabcolsep}{2pt}
\begin{tabular}{|ccc|l|}
\hline
\bf Level     &\bf  Grade      & \bf Age   & \multicolumn{1}{c|}{\bf Examples}  \\\hline\hline
\bf  I   &  1    & 6     &\multicolumn{1}{r|}{\<بَيْت، كَبير، أكَلَ، عَلى>}\\
&&&  house, big, to eat, on\\\hline
\bf  II  &  2-3 & 7-8   & \multicolumn{1}{r|}{\< ذَهَب، أُسْطُواني، خَدَعَ، إذا >}\\
&&&  gold, cylindrical, to cheat, if\\\hline
\bf  III &  4-5  & 9-10  & \multicolumn{1}{r|}{\< رِئة، مُعادَلة، مُوَحَّد، أَغْرى>}\\
 &&&  lung, equation, united, to entice \\\hline
\bf  IV  &  6-8 & 11-14 &\multicolumn{1}{r|}{ \<اِقْتِصاد، طُمَأنينة، راقِي، نَكَثَ>    }\\
&&&   economy, tranquility,  \\
&&&   sophisticated, to breach \\\hline

\bf  V   & 9+ & 15 -  & \multicolumn{1}{r|}{\<أَدَمة، مِطْياف، لَوْذَع، شُعَبيّ>}  \\
&&&   epidermis, spectroscope,\\ 
&&&   witty, bronchial\\ \hline
\end{tabular}
\caption{The five readability levels, their grade equivalencies, and lemma and English gloss examples, abridged from \newcite{al-khalil-etal-2020-large}. }
\label{tab:readabilitylevels}
\end{table}

\section{Related Work}
\label{related}
\paragraph{Readability Resources} Text readability leveling is relevant to a wide range of NLP applications such as text simplification and automatic readability assessment. Most research on readability leveling has focused on English, leading to the development of many resources \cite{collins-thompson-callan-2004-language,pitler-nenkova-2008-revisiting,feng-etal-2010-comparison,vajjala-meurers-2012-improving,xia-etal-2016-text,nadeem-ostendorf-2018-estimating,vajjala-lucic-2018-onestopenglish,deutsch-etal-2020-linguistic,lee-etal-2021-pushing}. 
%Although other languages, particularly morphologically rich ones, have not received as much attention, there have been some research efforts on readability assessment and leveling in terms of publicly available corpora and tools. 

Specifically for MSA, datasets and modeling approaches have been created and developed by leveraging text targeted towards L1 readers (native speakers) \cite{Al-Khalifa:2010:automatic,AlTamimi:2014:aari,el-haj-rayson-2016-osman,Khalil:2018:leveled} and L2 learners (non-native speakers) \cite{Forsyth:2014:automatic,Saddiki:2018:feature}. More recently, \newcite{al-khalil-etal-2020-large} developed a 26,578-lemma lexicon (later extended to over 40,000 lemmas) 
%get exact numbers!
with a five-level readability scale. Examples of vocabulary from the different readability levels and their corresponding grades and ages are shown in Table~\ref{tab:readabilitylevels}. This lexicon anchors readability at the lemma representation of the  words.
%Level I (Grade 1, age 6), Level II (Grade 2-3, age 7-8), Level III (Grade 4-5, age 9-10), Level IV (Grade 6-8, age 11-14), Level V (specialist, age 15 and above). 
We use this lexicon as our reference for readability levels.

\newcite{Jiang:2020:online} developed the online Readability Leveled Arabic Thesaurus interface that leverages  \newcite{al-khalil-etal-2020-large}'s  lexicon, and extends its coverage.\footnote{\url{http://samer.camel-lab.com/}} For a
given user input word, this interface provides the word’s possible lemmas, roots, English glosses, related Arabic words and phrases from the Arabic WordNet \cite{Black:2006:introducing}, and readability on a five-level readability scale.
We make use of many components of \newcite{Jiang:2020:online}'s interface  in our add-on.

%Notes from Reem :
% Re (1), I found a few mismatches:
% 
% The word “كَبير” has a gloss of “great, important, large, major, % senior”, but not “big”. 
% 
% The word “ذهب” has a readability level of 1, not 2.
% 
% The word “خَدَعَ” has a gloss of “deceive”, not “to cheat”.
% 
% The word “إذا” has a readability level of 1, not 2.
% 
% The word “رِئة” has a readability level of 2, not 3. 
% 
% The word “مُعادَلة” has a readability level of 4 and the gloss % “balancing, equalizing”.
% 
% The word “ أَغْرى” has a gloss of "be seduced, incite, induce, % provoke”, but not “to entice”
% 
% The words “اِقْتِصاد” , “راقِي”  and “طُمَأنينة” have a readability % level of 3, not 4.
% 
% The word “نَكَثَ” has a gloss of “violate”, not breach.
% 
% The word أَدَمة has a gloss of “skin”.

\paragraph{Readability Visualization} To the best of our knowledge, there has not been much work on developing web-based visualization tools for word-level readability assessment, neither for Arabic nor for other languages. Most of the existing tools work on the document or the sentence levels. Such tools include Readable\footnote{\url{https://readable.com/}} and datayze's Readability Analyzer\footnote{\url{https://datayze.com/readability-analyzer.php}} for English, and the recently proposed FABRA for French~\cite{wilkens-EtAl:2022:LREC}.\footnote{\url{https://cental.uclouvain.be/fabra/}} The lack of word-level tools for Arabic has motivated us to create an easy-to-use Google Docs add-on for 
%contextualized 
word-level readability visualization. 
% https://github.com/CAMeL-Lab/samer-python-api 

%\paragraph{Arabic Readability Interfaces} 
%Two particular interfaces 
%designed to assist speakers and learners in understanding Arabic written text have inspired some of our design decision.  First, \newcite{khalifa-etal-2016-dalila} presented a Chrome extension that utilizes various Arabic NLP tools to assist learners and non-native speakers in understanding text written in either MSA or dialectal Arabic (DA). 
%And secondly, \newcite{Jiang:2020:online} developed the online Readability Leveled Arabic Thesaurus interface that leverages  \newcite{al-khalil-etal-2020-large}'s  lexicon, and extends its coverage.\footnote{\url{http://samer.camel-lab.com/}} For a given user input word, this interface provides the word’s possible lemmas, roots, English glosses, related Arabic words and phrases from the Arabic WordNet \cite{Black:2006:introducing}, and readability on a five-level readability scale.

\paragraph{Arabic Morphological Analysis and Disambiguation} There are a number of tools that support Arabic morphological analysis and disambiguation and specifically lemmatization \cite{Pasha:2014:madamira, Darwish:2016:farasa, obeid-etal-2020-camel,Obeid:2022:Camelira}.
Inspired by the JavaScript Chrome extension developed by  \newcite{khalifa-etal-2016-dalila}
to assist Arabic learners in understanding text written in MSA or dialectal Arabic (DA),
 we 
%use the Camel~Tools library \cite{obeid-etal-2020-camel} in our API, and we 
implement a version of the Buckwalter core morphological analysis algorithm \cite{Buckwalter:2002:buckwalter} in JavaScript as part of our add-on.

\section{Design and Implementation}
\label{design}

\subsection{Design Considerations}

We designed our interface %and supporting APIs 
with the following considerations in mind.

\paragraph{Openness and Ease-of-use} 
The system needs to be powerful and provides additive or complementary value to existing text editors, so that simplifications and changes can be evaluated on the fly and with minimal overhead. This needs to be accomplished with minimal usability tradeoffs.

\paragraph{Handling Arabic Ambiguity and Rich Morphology} 
The system needs to be able to analyze fully inflected words and relate them to their lemmas and part-of-speech (POS) tags. The lemmas and POS tags will be used to identify the readability levels from \newcite{al-khalil-etal-2020-large}'s lexicon and to link with the Arabic WordNet databases \cite{Black:2006:introducing}.
Additionally, the interface needs to provide the users with access to all the analyses of a given word.

\paragraph{Visualizing Readability} 
The interface needs to provide
summary readability statistics in word-token and word-type spaces over
full documents or arbitrary text selections.  It should highlight the words in context in a clear way to indicate intuitively which words are easier and which are harder. And finally, the interface needs to provide access to the readability levels of other unchosen analyses of any word.  
%We will use \newcite{al-khalil-etal-2020-large}'s readability levels as our reference. %But other approaches could also be used.

\paragraph{Access to Word Substitutions} 
The system should support the text simplification process by displaying suggestions for related words and phrases, e.g., synonyms, antonyms, hypernyms, and hyponyms, with different readability levels.
 We build on the work of \newcite{Jiang:2020:online} who used the Arabic Wordnet to accomplish the same.

 \begin{figure*}[t!]
\centering
    \includegraphics[width=1.8\columnwidth]{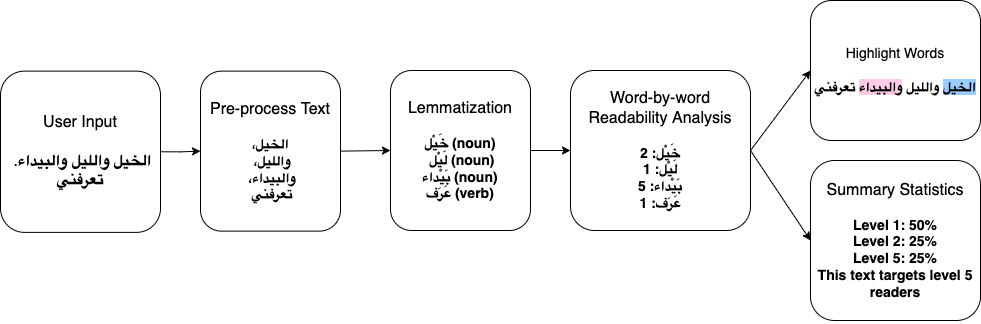}
    \caption{A flowchart depicting the steps that our tool takes to process user input. %As an example of user input, we use a classic verse from the famous Arab poet, Al-Mutanabbi. 
    First, the user input is pre-processed and tokenized. Next, the lemma and part-of-speech of each token are determined using a morphological analyzer. Then, the tool looks up each lemma in the readability database to identify its readability level. The tool then highlights individual words accordingly and produces summary statistics describing the overall text readability. }
    \label{fig:flowchart}
\end{figure*}

\paragraph{Explicit/Implicit Word Readability Markup} 
The system should allow the recording of explicit readability levels such that when
the automatic processes make mistakes, users can overwrite them. We want those corrections and annotations to be persistent across different future versions of the analyzer and lexicon. At the same time, unnecessary over-specification can be distracting to the reader or annotator and should be minimized. The system  should support the ability to import and export text files that could be marked for readability using external tools.

\subsection{Implementation}
%\label{implementation}

\paragraph{Google Docs Add-on}
We opted to implement our interface as a Google Docs add-on, which allows us to use one of the world's most used editing frameworks, without sacrificing any of Google Docs' advantages such as multi-author editing and other familiar word-editing supports.

We implemented the tool's front-end in HTML, CSS and JavaScript. The back-end was implemented in JavaScript, and it also utilizes the Apps Script Document Service, which is a JavaScript API used to read and modify Google Docs programmatically. 
%We link the front-end and back-end using JavaScript event handlers, which detect user clicks in the frontend and trigger the backend process.

%We plan to use two different disambiguation approaches, one computationally light but less accurate, and one more expensive and accurate.  

%\paragraph{Morphological Resources}

% https://developers.google.com/apps-script/reference/document

% link to figure
% https://drive.google.com/file/d/1dAOaOUT87oIxhi7dq3SRM5xM1_yDixNy/view?usp=sharing
% import into draw.io

\paragraph{Readability Analysis and Visualization}

The tool analyzes user input in four main steps that are summarized in Figure~\ref{fig:flowchart}. First, in the back-end, the text is pre-processed and tokenized, and non-word tokens are discarded. Second, the tokens are fed into the  morphological analysis algorithm, which produces the most likely lemma and POS pair for each word. Third, we look up the lemmas in the readability database to identify their readability levels.\footnote{We treat Proper nouns (Names) as a separate level.} Finally, we use the Apps Script Document API to highlight words with different colors according to their readability levels. The tool also presents a summary of the text's readability distribution levels in a bar chart  colored consistently with the readability level word highlights. 

\paragraph{Morphological and Lexical Analyses}
Inspired by \newcite{khalifa-etal-2016-dalila}'s Chrome extension and \newcite{obeid-etal-2020-camel}'s out-of-context MLE disambiguation mode, 
 we
implemented a version of the Buckwalter core morphological analysis algorithm \cite{Buckwalter:2002:buckwalter} in JavaScript as part of our add-on.
Besides being used to determine readability levels, all lemma analyses are presented in a side bar to allow investigating and reassigning  readability levels if needed.
It is worth noting that the readability lexicon we use does not handle lexical polysemy. This is mainly due to the lexical representation that is used in the lexicon, which follows the representation of the Standard Arabic Morphological Analyzer (SAMA) \cite{Graff:2009:standard}. However, the design of our tool is independent of the granularity level of lexical representation and therefore, any updates to these components in the future can be easily integrated in our tool.

\begin{figure*}[t]
\centering
    \frame{\includegraphics[width=1.9\columnwidth]{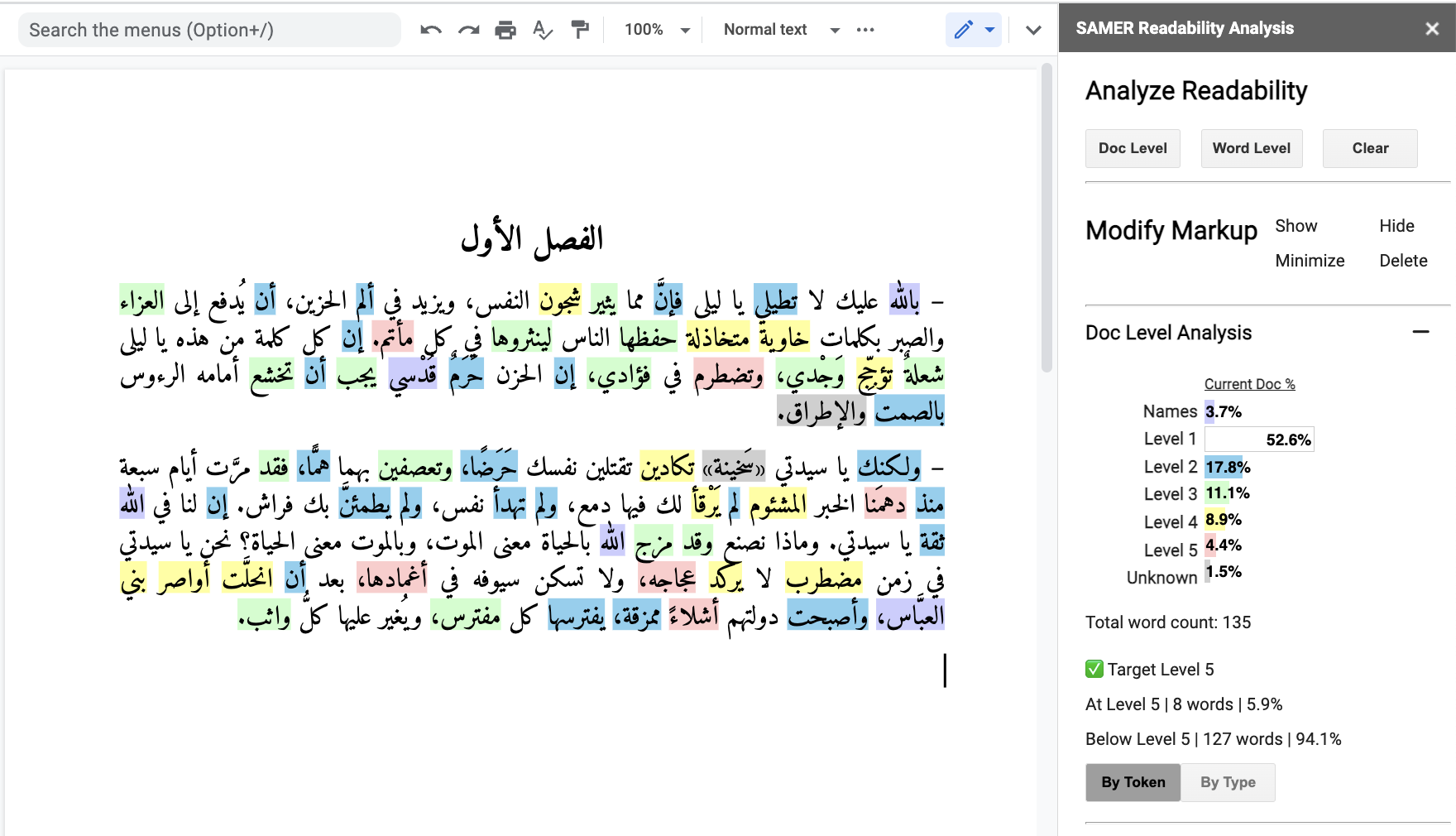}}
    \caption{The SAMER Google Docs add-on visualizing  word-level and document-level readability.}\label{fig:doc-level}
\end{figure*}

Figure~\ref{fig:doc-level} presents an instance of the SAMER Google Docs add-on with marked up text.

\begin{figure}[t]
\centering
    \includegraphics[width=0.9\columnwidth]{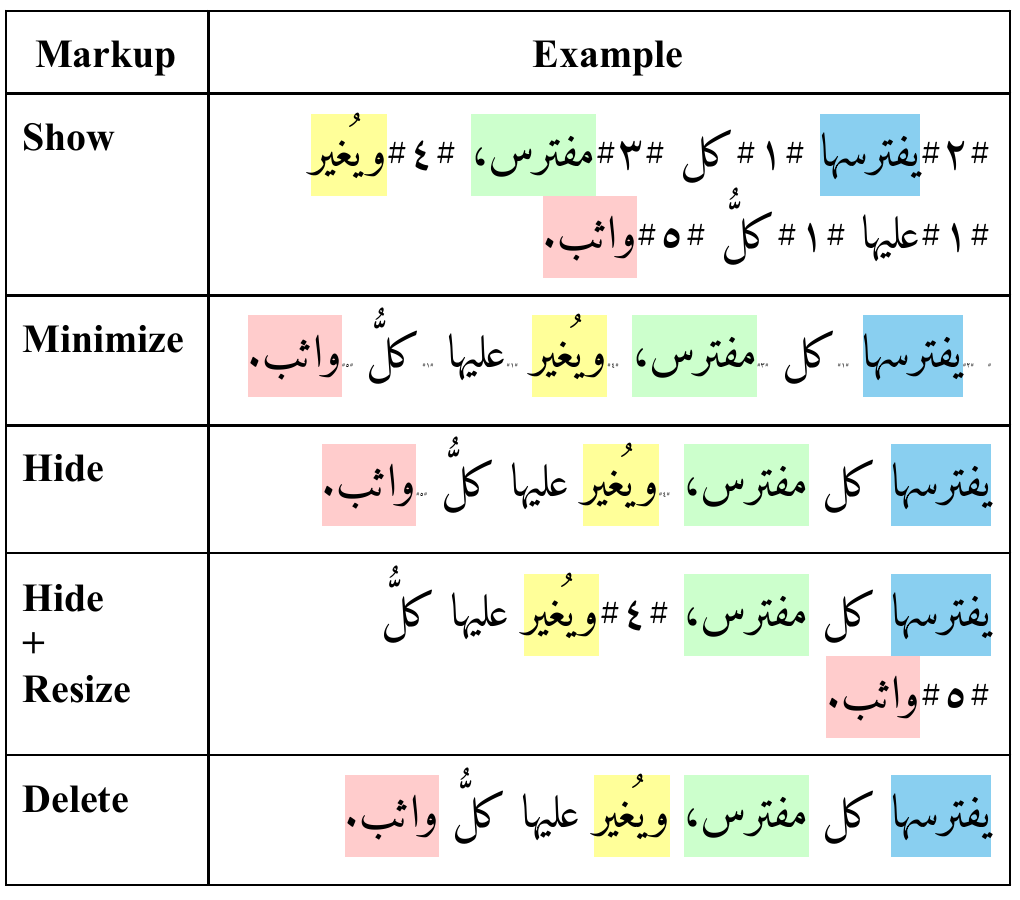}
    \caption{The different word-level markup modes that are supported by our tool.}
    \label{fig:markups}
\end{figure}

\paragraph{Explicit/Implicit Word Readability Markup}
By default, the system deterministically specifies a readability level for any specific word based on its morphological and lexical readability resources.  When disagreement with the automatic levels happen, as in automatic errors or importing text that was leveled externally,  we ensure that the differences from the deterministic readability levels are not lost.  To accomplish this,  a prefix \textit{\#<i>\#} is explicitly added in front of the word in question forcing the tool to interpret the word as having  readability level of value  \textit{<i>}. For example, the word \<كتب> {\it ktb} has a readability level of 1. However, the user can manually assign it a level of 5 by adding \#\<٥>\# (Indo-Arabic digit~{\it 5}) in front of the word, like so: \<كتب>\#\<٥>\# {\it \#5\#ktb} .  We also provide an interface button as part of the morphological side bar discussed above to make such assignment.

The add-on also provides multiple markup visualization modes to navigate between explicit and implicit readability level markup.
\begin{enumerate}[label=(\alph*)]
    \item \textbf{Show}: Explicitly mark all words with their readability levels.
    %by adding \textit{\#<i>\#} in front of each word.
    \item \textbf{Minimize}: Minimize all the markups by setting their font size to 1pt.
    \item \textbf{Hide}: Remove any markup whose readability level matches the internal level chosen by the analyzer and  only keep the disagreeing markups. 
    %This option cleans up the text while only keeping the necessary markup. 
    By default the Hide mode also minimizes the markup; however, the user can easily select the full text and resize it to a preferred font size (Hide+Resize).
    \item \textbf{Delete}: Delete all markup from the text.
\end{enumerate}

Figure~\ref{fig:markups} shows the supported markup modes. % supported by our tool. %We discuss all of the above implementation decisions with examples in the next section.

\begin{figure*}[t]
\centering
    \frame{\includegraphics[width=1.9\columnwidth]{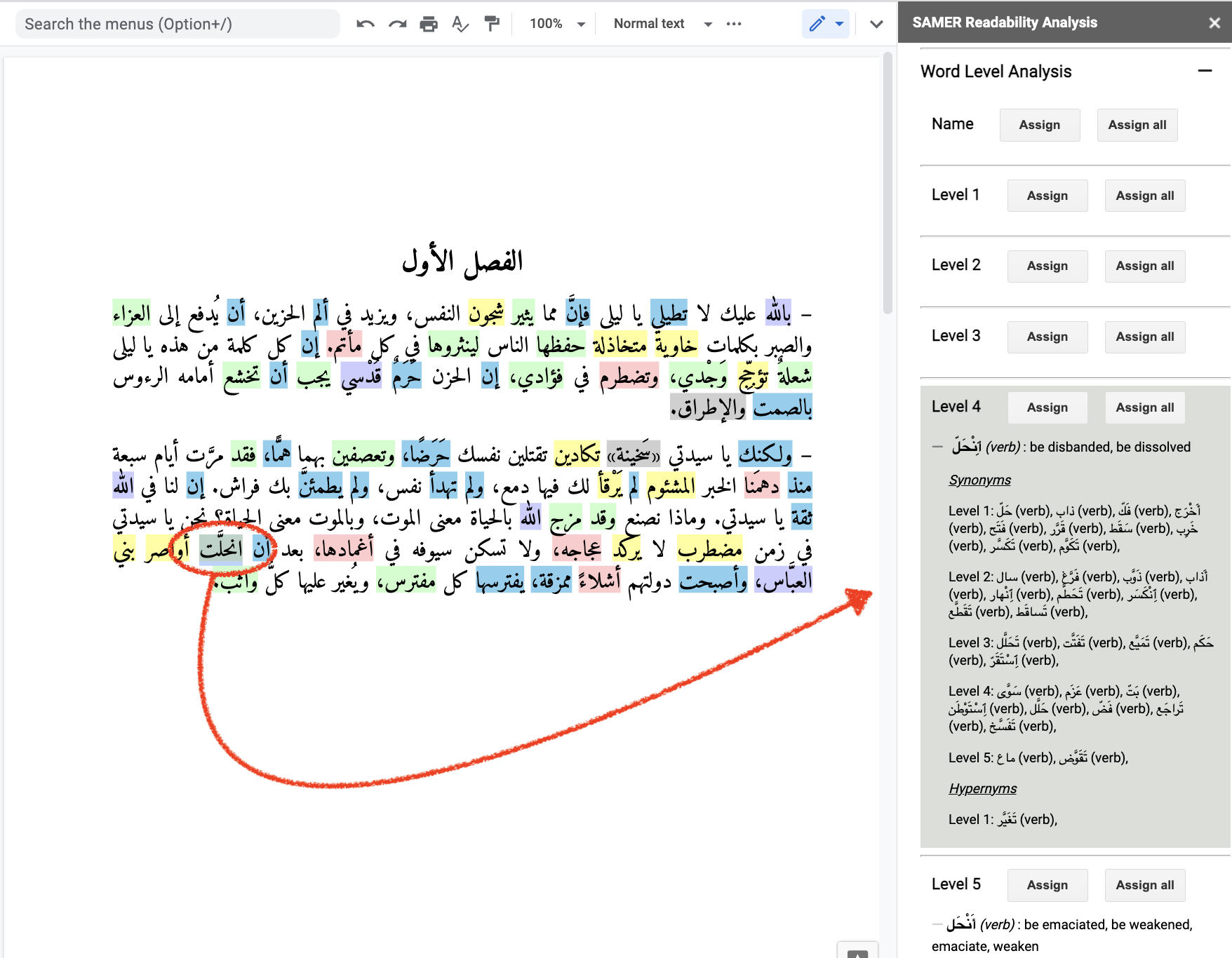}}
    \caption{An example of selecting a specific word and identifying all of its analyses with their readability levels.}\label{fig:word-level}
\end{figure*}

% **************ORIGINAL TEXT EXAMPLE*********************
% *     https://www.hindawi.org/books/93085140/8/        *
% ********************************************************
\section{SAMER Add-on: Examples and Use Cases}
\label{example}

We present some examples of how the SAMER project Google Docs add-on can be used to analyze the readability of a literary text. We also discuss potential use cases of our tool across a variety of tasks and how it can be extended to other languages.
\paragraph{Examples} Figure~\ref{fig:doc-level} shows the result of using the tool to analyze a short segment of a novel. After clicking on the \texttt{Doc Level} button at the top, the tool highlights each word according to its readability level using different colors, and presents a summary distribution of words in each readability level.% and displays the number of words and overall readability level of the text.

Figure~\ref{fig:word-level} shows the result of selecting a specific word (\<انحلت> {\it AnHlt} `be disbanded') and clicking on the \texttt{Word Level} button at the top. A side bar appears showing the different lemma analyses by readability level. Various word substitution alternatives are presented to the user including synonyms, hypernyms and hyponyms, with their associated readability levels.  If the user decides to change the word, they can simply rewrite it and rerun the readability analysis. If the user decides to change the automatically assigned readability level, they can either change it directly manually, or by clicking on the \texttt{Assign} button to change that specific word's readability level markup or the \texttt{Assign All} button to change all of its occurrences in the document.

\paragraph{Use Cases} Our goal behind creating an easy-to-use Google Docs add-on tool for Arabic word-level readability analysis is to enable users to edit texts easily based on a specific target  readability level. We intend for our tool to be used by human annotators to identify text readability levels and to simplify texts in a controlled setting. However, we envision that our tool can be used to assist writers in either making texts more sophisticated (harder readability) or in providing alternatives for specific words that have the same readability level.

\paragraph{Extending to Other Languages} Although our work focuses on Arabic, the SAMER add-on tool is designed in a modular way and it can be easily extended to other languages. More concretely, the following core components are needed to make such an extension possible: (1) a readability level lexicon that relates lemmas to their readability levels; (2) a morphological analysis database that specifies prefixes, suffixes, stems and lemmas, and their co-occurrence compatibilities; (3) a statistical lemma-based disambiguation model; and (4) synonym, hypernym, hyponym and antonym lexical databases,  such as those found in  WordNet \cite{Fellbaum:2010:wordnet}. 
%\begin{figure*}[t]
%\centering
%    \includegraphics[width=1.8\columnwidth]{figures/Orig-simpl.pdf}
%    \label{fig:related-words}
%\end{figure*}

% Figure~\ref{fig:markups} shows the various markup modes supported by our tools.

\section{Conclusion}

We presented a Google Docs add-on for automatic Arabic word-level readability visualization. Our add-on includes a lemmatization component that is connected to a five-level readability lexicon and Arabic WordNet-based substitution suggestions. The add-on can be used for assessing the reading difficulty of a text and identifying difficult words as part of the task of manual text simplification. 

In future work, we plan on enhancing our tool's readability analysis by leveraging additional morphosyntatic features \cite{Saddiki:2018:feature}. We will use the add-on to annotate
a corpus of parallel original and simplified texts from Arabic novels.
% this work could be used as annotation interface to create data for Arabic simplification

% We are planning large efforts 

\section*{Limitations and Ethical Considerations}
% cheating; profiling; errors; lemmas

We acknowledge that the add-on we developed could be used maliciously to:
(a) modify texts under false pretenses, (b) plagiarize, or (c)
profile people in a biased way using their writing style. 
We also acknowledge that automatic errors in readability analysis can lead to harmful results even when used with good intent.
We further recognize that the use of highlighting as a visualization mechanism limits the conventional use of highlighting in text editing.
Another limitation of our work is the lack of extrinsic and intrinsic evaluation.
%when it comes to assigning the readability levels to the words. 
However, we are not aware of any manually annotated Arabic word-level readability  datasets .
We plan to develop such datasets using our tool.
Finally, we acknowledge that further user studies are needed to confirm the effectiveness of our tool in aiding annotators to perform tasks such as text simplification.

\section*{Acknowledgements}
This project is funded by a New York University Abu Dhabi Research Enhancement Fund grant. We thank Zhengyang Jiang,  Go Inoue, and Ossama Obeid for  helpful discussions.

\bibliography{anthology,custom,camel-bib-v2}
\bibliographystyle{acl_natbib}

% \appendix

% \section{Example Appendix}
% \label{sec:appendix}

\end{document}